\documentclass[10pt,twocolumn,letterpaper]{article}

\usepackage{cvpr}
\usepackage{times}
\usepackage{epsfig}
\usepackage{graphicx}
\usepackage{amsmath}
\usepackage{amssymb}
\usepackage{caption}  
\usepackage{subcaption}
\usepackage[pagebackref=true,breaklinks=true,letterpaper=true,colorlinks,bookmarks=false]{hyperref}

\cvprfinalcopy 

\def\cvprPaperID{72} 

\ifcvprfinal\fi
\begin{document}

\title{EgoTransfer: Transferring Motion to First Person Vision}


\title{EgoTransfer: Transferring Motion Across Egocentric and Exocentric Domains using Deep Neural Networks}

\author{Shervin Ardeshir\\
{\tt\small ardeshir@cs.ucf.edu},\\
\and
Krishna Regmi\\
{\tt\small krishna.regmi7@gmail.com}\\
\and
Ali Borji\\
{\tt\small aborji@crcv.ucf.edu}\\
\and
Center for Research in Computer Vision (CRCV)\\
University of Central Florida, Orlando, FL\\
}

\maketitle

\begin{abstract}
Mirror neurons have been observed in the primary motor cortex of primate species, in particular in humans and monkeys. A mirror neuron fires when a person performs a certain action, and also when he observes the same action being performed by another person. A crucial step towards building fully autonomous intelligent systems with human-like learning abilities, is the capability in modeling the mirror neuron. On one hand, the abundance of egocentric cameras in the past few years has offered the opportunity to study a lot of vision problems from the first-person perspective. A great deal of interesting research has been done during the past few years, trying to explore various computer vision tasks from the perspective of the self. On the other hand, videos recorded by traditional static cameras, capture humans performing different actions from an exocentric third-person perspective. In this work, we take the first step towards relating motion information across these two perspectives. We train models that predict motion in an egocentric view, by observing it from an exocentric view, and vice versa. This allows models to predict how an egocentric motion would look like from outside. To do so, we train linear and nonlinear models and evaluate their performance in terms of retrieving the egocentric (exocentric) motion features, while having access to an exocentric (egocentric) motion feature. Our experimental results demonstrate that motion information can be successfully  transferred across the two views.

\end{abstract}

\section{Introduction}

According to Wikipedia:\\
\textit{"A mirror neuron is a neuron that fires both when an animal acts and when the animal observes the same action performed by another. Thus, the neuron mirrors the behavior of the other, as though the observer were itself acting. Such neurons have been directly observed in primate species."}\\

\begin{figure}[h]
\centering
\includegraphics[width=1\linewidth]{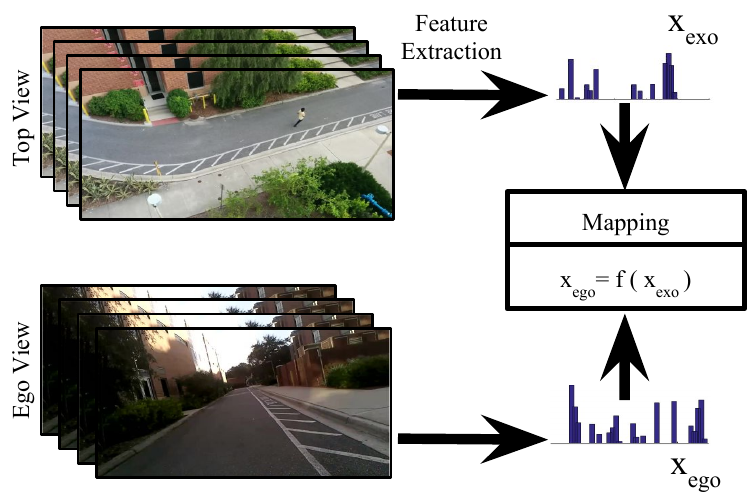}
\caption{The main objective of this work is to learn a mapping between egocentric and exocentric motion features. Having a set of time-synchronized egocentric-exocentric pairs of videos, we aim to extract different types of motion features from both views and train different mapping models to automatically learn their relationship.}
\label{fig:blockDiagram}
\vspace{-10pt}
\end{figure}

Achieving human-like learning abilities, requires modeling the mirror neuron phenomenon. In other words, intelligent systems should be able to relate visual information from a third person perspective, to first person perspective, and vice versa. Watching a human running, an intelligent system (e.g., a robot) should be able to imagine how the visual world would look like, if the system itself actually attempted such an act. We believe now is the perfect time for taking the first step towards modeling this concept from a computer vision standpoint.

During the past few years, egocentric cameras have provided us the opportunity to study first person vision widely and extensively. Thanks to the affordability of wearable cameras and smart glasses (e.g., GoPro, Google glass), a lot of interesting research has been done ranging from action recognition \cite{egoDailyAction} to identification and localization\cite{ardeshir2016ego2top}.
The history of computer vision, however, goes beyond the past few years. Tremendous amount of research has been conducted in different areas of computer vision but on more traditional types of videos collected using static cameras from canonical, oblique or top view. We refer to these videos as exocentric or third-person videos. Given the fact that egocentric vision is a relatively new area , compared to exocentric vision, the amount of available exocentric data is drastically more than egocentric data. For example, while there are several datasets in the computer vision community for action and activity recognition in exocentric domain, there are not nearly as many egocentric datasets.

In order to take advantage of the vast amount of knowledge that exists in the exocentric domain, there is a need for a systematic adaptation of exocentric information to the egocentric domain. In this study, we explore the relationship between egocentric and exocentric for visual motion transfer. In other words, we seek to learn a transformation from motion features in exocentric space, to that in egocentric space and vice versa. To do so, we collect a dataset of egocentric and exocentric videos captured simultaneously with body mounted egocentric and static exocentric cameras, capturing people performing diverse actions covering a broad spectrum of motions. We then divide each pair of videos to time-synchronized short clips of 16 frames, and extract motion features from each clip in each view as illustrated in Figure \ref{fig:blockDiagram}. This will provide us a set of feature pairs $(x_{exo},x_{ego})$ from the two views. We then train different linear and non-linear mappings to learn a transformation between the two views. For testing the performance of our models, we evaluate their capability in terms of retrieving their correct match from the other view. In the test set, we have a set of video pairs, and therefore feature pairs, extracted from simultaneously recorded ego and exocentric videos. We map each feature from the first view (source view) to the other view (target view), and evaluate its capability in terms of finding its correct paired video/feature in the target set. In other words, we try to retrieve it's corresponding video clip in the test set. We evaluate and analyze the performance of different mapping methods in different scenarios. 

\begin{figure}
\centering
\includegraphics[width=1\linewidth]{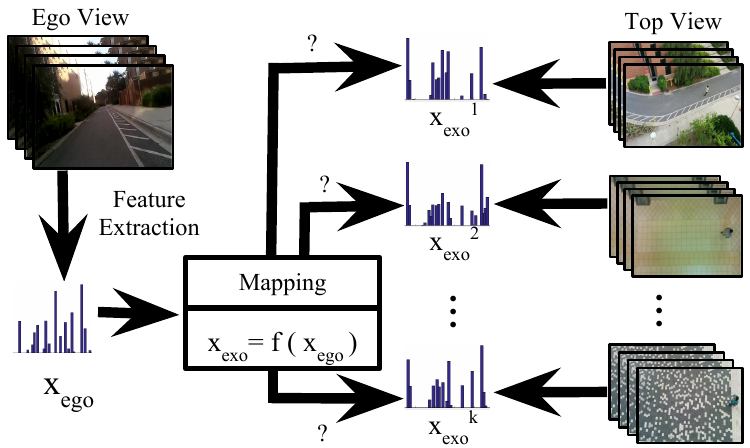}
\caption{The testing process for evaluating our models. Having a clip in the one view (here egocentric), we first extract its motion features. We then apply our models to transform the egocentric motion features to the exocentric space. The set of exocentric videos are ranked based on their feature similarity to the transformed egocentric features and the rank of the correct ground-truth exocentric video is used for evaluating the performance. We compare our  models in terms of their ranking capability.}
\label{fig:retrievalProcess}
\vspace{-10pt}
\end{figure}

\section{Related Work}
First person vision, also known as egocentric vision, has became increasingly popular in the vision community. A lot of research has been conducted in the past few years ~\cite{egoKanade,egoEvolutionSurvey}, including object detection \cite{egoObjectDetection}, activity recognition \cite{egoDailyAction,egoActionFathi} and video summarization \cite{egoVideoSummarization}.

Motion in egocentric vision, in particular, has been studied as one of the fundamental features of first person video analysis. Costante et al. \cite{costante2016exploring} explore the use of convolutional neural networks (CNNs) to learn the best visual features and predict the camera motion in egocentric videos. Su and Grauman \cite{DBLP:journals/corr/SuG16a} propose a learning-based approach to detect user engagement by using long-term egomotion cues. Jayaraman et al. \cite{DBLP:journals/corr/JayaramanG15} learn the feature mapping from pixels in a video frame to a space that is equivariant to various motion classes. Ma et al. \cite{DBLP:journals/corr/MaFK16} have proposed a twin stream network architecture to analyze the appearance information and the motion information from egocentric videos and have used these features to recognize egocentric activities.

Action and activity recognition in egocentric videos have been hot topics in the community. Ogaki et al. \cite{6239188} jointly used eye motion and ego motion to compute a sequence of global optical flow from egocentric videos. Poleg et al. \cite{DBLP:journals/corr/PolegEP015} proposed a compact 3D Convolutional Neural Network (3DCNN) architecture for long-term activity recognition in egocentric videos and extended it to egocentric video segmentation. Singh et al. \cite{Singh_2016_CVPR} used CNNs for end-to-end learning and classification of actions by using hand pose, head motion and saliency map. Li et al. \cite{Li_2015_CVPR} used gaze information, in addition to these features, to perform action recognition.
In their work, Matsuo et al. \cite{Matsuo_2014_CVPR_Workshops} have proposed an attention based approach for activity recognition by detecting visually salient objects.
The relationship between egocentric and top-view videos has been explored in tasks such as human identification \cite{ardeshir2016ego2top,ardeshiregocentric}  and temporal correspondence\cite{}.
The relationship between egocentric and top-view information has been explored in tasks such as human identification \cite{ardeshir2016ego2top,ardeshiregocentric}, semantic segmentation\cite{ardeshir2015geo}  and temporal correspondence\cite{ardeshirEgocentricMeets}. In this work, we relate two different views of a motion, which can be considered as a knowledge transfer or domain adaptation task. Knowledge transfer has been used for the multi-view action recognition (e.g., \cite{junejo2008cross,liu2011cross,li2012discriminative}) in which multiple exocentric views of an action are related to each other. Having multiple exocentric views allows geometrical and visual reasoning, since: a) the nature of the data is the same in different views and b) the actor is visible in all cameras. In contrast, our paper aims to automatically learn mappings between two drastically different views, egocentric and exocentric. To the best of our knowledge, this is the first attempt in relating these two domains for transferring motion information.

\section{Framework}
Our main goal is to transform motion features from a \textit{source} view to a \textit{target} view, where one is egocentric and the other is exocentric. In other words, we try to learn mapping models from egocentric to exocentric space, and vice versa. As shown in figure \ref{fig:blockDiagram}, we have a set of datapoint pairs $(\mathbf{x}_{source},\mathbf{x}_{target})$, in this case $(\mathbf{x}_{ego},\mathbf{x}_{exo})$, in the training set for which we try to learn a transformation that maps one view to another i.e. estimates a mapping function $f$, such that $\mathbf{x}_{ego}=f(\mathbf{x}_{exo})$ or $\mathbf{x}_{exo}=f(\mathbf{x}_{ego})$. The datapoints are in fact spatiotemporal features such as C3D (3D neural network based spatiotemporal features \cite{c3dtran2015learning}), and HOOF features  \cite{chaudhry2009histograms} capturing histogram of oriented optical flow. We then train linear and non-linear mapping models using these pairs. We evaluate the learned models one a test set, in terms of their capability in retrieving the groundtruth paired feature from the other view. As shown in figure \ref{fig:retrievalProcess}, for each ego/exocentric video, we extract its motion feature descriptor and retrieve its correct paired video from the other view. We then rank the videos in the other view. The rank of the correct match will be a metric for us to evaluate different models over different scenarios.  

\subsection{Extracting Motion Features} We represent each short video clip in each view using its motion feature. Two different motion features with different levels of complexity are employed. We use the simple feature of histogram of oriented optical flow (HOOF), and also a more complicated spatiotemporal feature known as 3D convolutional neural networks (C3D) proposed in \cite{c3dtran2015learning}. We study the mapping capacity of these features from/to egocentric videos to/from exocentric videos (top-view and side-view) using different mapping methods.\\

\noindent\textbf{C3D Features:} These 4096D feature descriptors are computed using a 3D convolutional neural network \cite{c3dtran2015learning}. In order to reduce the computational complexity in training our models, we reduce the dimensionality to 128D using Principal Component Analysis (PCA).\\

\noindent\textbf{HOOF Features:} Histogram of Oriented Optical Flow was extracted with 32 different orientations, resulting in a 32D feature vector representing each clip.\\

\subsection{Mapping Models:} We train 3 linear baseline models and two different non-linear models to evaluate the possibility of learning a mapping between egocentric and exocentric motion features. In what follows, we explain the details of each mapping scheme alongside with the implementation details.

\subsubsection{Linear Models:} We train 3 different linear models of uniform transformation (direct matching), linear regression, and linear regression with L2 regularization as baselines.

\noindent\textbf{Direct Matching:} One might ask the question that how the feature descriptors would perform if they were directly compared. In other words, what if we simply retrieve the exocentric videos, by directly comparing them with the egocentric query feature. To answer whether complicated mappings are necessary, we evaluate the performance of direct matching between the two spaces. Direct matching assumes a uniform transformation across the two spaces.
Given that source and target domains are totally different, direct matching is not expected to outperform chance significantly. Our experiments also validate this expectation, as direct matching always achieves near-random performances. 

\noindent\textbf{Linear Regression:} We tried training a linear regression model from the source domain to the target domain. Linear regression has the following form and can be computed using a closed form solution with least squares optimization.  
\begin{equation}
\underset{\textbf{w}}{\operatorname{argmin}} ||\textbf{w}\mathbf{x}_{source} - \mathbf{x}_{target}||^2 
\end{equation}
Our experiments indicate that linear regression consistently outperforms chance and direct matching with a large margin, but it suffers from the limits of linear models. This consistent edge compared to direct matching, suggests that better mappings are possible. 

\noindent\textbf{Regularized Linear Regression:} Regularization has shown to improve regression models by preventing them from overfitting to data and converging to trivial solutions. We tried L2 regularized linear regression models and evaluated their accuracies as well. In our experiments, L2 regularization does not improve the accuracy considerably compared to linear regression in most of the scenarios.

\subsubsection{Non-linear Models:} We test the performance of two generic neural network based architectures in order to explore the possibility of improving our linear baselines. Details are described in the following.\\

\begin{figure}[t]
\centering
\includegraphics[width=1\linewidth]{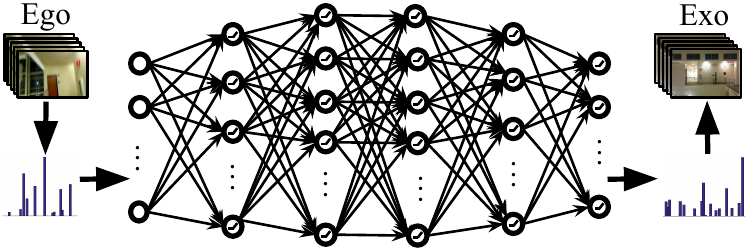}
\caption{The non-linear mapping model with fully connected layers, relu activation and batch normalization. This network tries to reconstruct the features in one view (ego/exocentric) from another view(exo/egocentric).}
\label{fig:nonLinear}
\vspace{-14pt}
\end{figure}

\noindent\textbf{Non-linear Mapping with a Reconstruction Objective:} We train a non-linear mapping model containing 5 layers of fully connected layers with relu activations and batch normalization, performed at each layer. Please see figure \ref{fig:nonLinear}. This architecture was designed for the purpose of reconstructing the target (ego/exocentric) features from the source features (exo/egocentric). We used the least square loss and adam optimizer for training. For HOOF features, the dimensionality of the fully connected layers are 32, 64, 128, 64, and 32. For C3D features, the dimensionalities are 128, 256, 256, 128, and 128. Our experiments show that this simple architecture is able to outperform the linear model with a large margin in most of the cases. Numbers of training and testing examples can be found in table \ref{tab:trainTest}. First, we train the model using batch size of 100 for 60 epochs. We then find the epoch number where the validation loss is minimum (i.e., the optimal number of epochs). Finally, we add the validation set to the training set, and train the model from scratch, up to the optimum number of epochs. We use the Keras platform with Theano backend 
to implement and test this network. \\ 

\begin{figure}[t]
\centering
\includegraphics[width=1\linewidth]{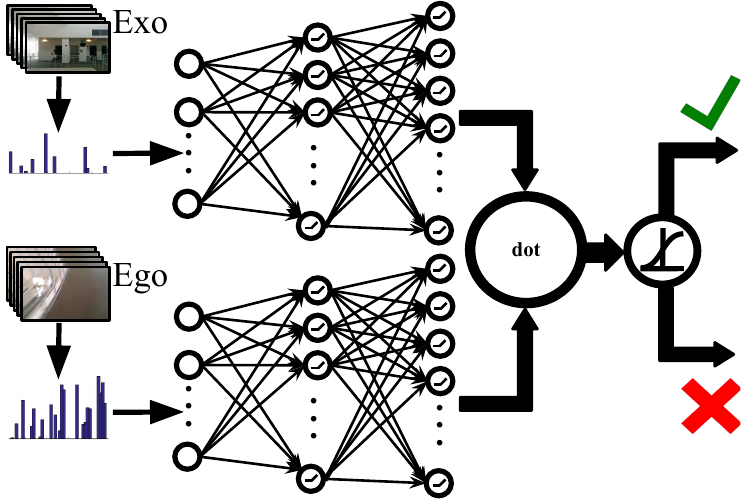}
\caption{The two-stream classification based network. The dense layer dimenionalities for HOOF are 64, and 128. And for C3D features the dimensionalities are 128 and 256. Again, the activations are relu and batch normalization is performed in each layer.}
\label{fig:twoStream}
\vspace{-12pt}
\end{figure}

\noindent\textbf{Two-stream Classification Network:} Two-stream classification networks have been one of the popular architectures for tasks such as matching and classification. We trained a two-stream neural network that requires pairs of features as input, and sends the dot product of the non-linear transformation of the two to a sigmoid function to enforce the notion of probability in the output. We choose the output to be 1 for corresponding feature pairs and 0 for non-corresponding pairs. The intuition behind this network is the assumption that there is a common space for the two views, to which they are non-linearly transformable, and therefore their dot product can be maximized. As shown in figure \ref{fig:twoStream}, we have two dense layers with relu activation and batch normalization applied to each stream. For the HOOF features, the dimensionalities of the dense layers in each stream are 64 and 128. For C3D features, the dimensionalities are 128 and 256. For training, positive and negative pairs are needed. Our original training data contains a lot of positive pairs. To generate negative pairs, we pick random non-correspondent features from the two views. Since the negative examples drastically outnumber the positive ones, we use sample weights in order to balance the training data. The weights for negative and positive examples, in order are $\frac{1}{n_{negative}}$ and $\frac{1}{n_{positive}}$. 

As in the other non-linear mapping method, we first train the model using batch size of 100 for 60 epochs. We then find the epoch number for which the validation loss is minimum. Finally, we add the validation set to the training set, and train the model from scratch and up to the optimum number of epochs. For optimizing this network, we use binary cross entropy loss.

\subsection{Testing our Mapping Models}
As shown in figure \ref{fig:retrievalProcess}, for each query clip in the source domain (egocentric in figure \ref{fig:retrievalProcess}) and in the test set, we aim to find its match in the target domain (exocentric in figure \ref{fig:retrievalProcess}). We extract the motion features of the query video. We then use our trained linear and non-linear models to transform the query source feature to the target domain. Finally, we compare the features extracted from the target set to the transformed query feature and rank them based on that. We rank target videos and evaluate the performance of the mapping models in terms of ranking the target videos. One would ideally want the correct match of the query video to appear at the top. For all the linear models and the non-linear mapping with reconstruction objective, this process is straightforward. For the two-stream network, we feed the query source feature, paired with each of the target features and acquire a score between 0 and 1, which denotes the probability of that target feature being matched with the query feature. Finally, we sort the target features in descending order, based on their score, and evaluate the method using the rank of the correct groundtruth match to the query feature.    

\section{Experimental Results}
In this section, we describe details of our collected dataset, evaluation metric, as well as performance of mapping models.

\begin{figure*}[t]
\centering
\includegraphics[width=1\linewidth]{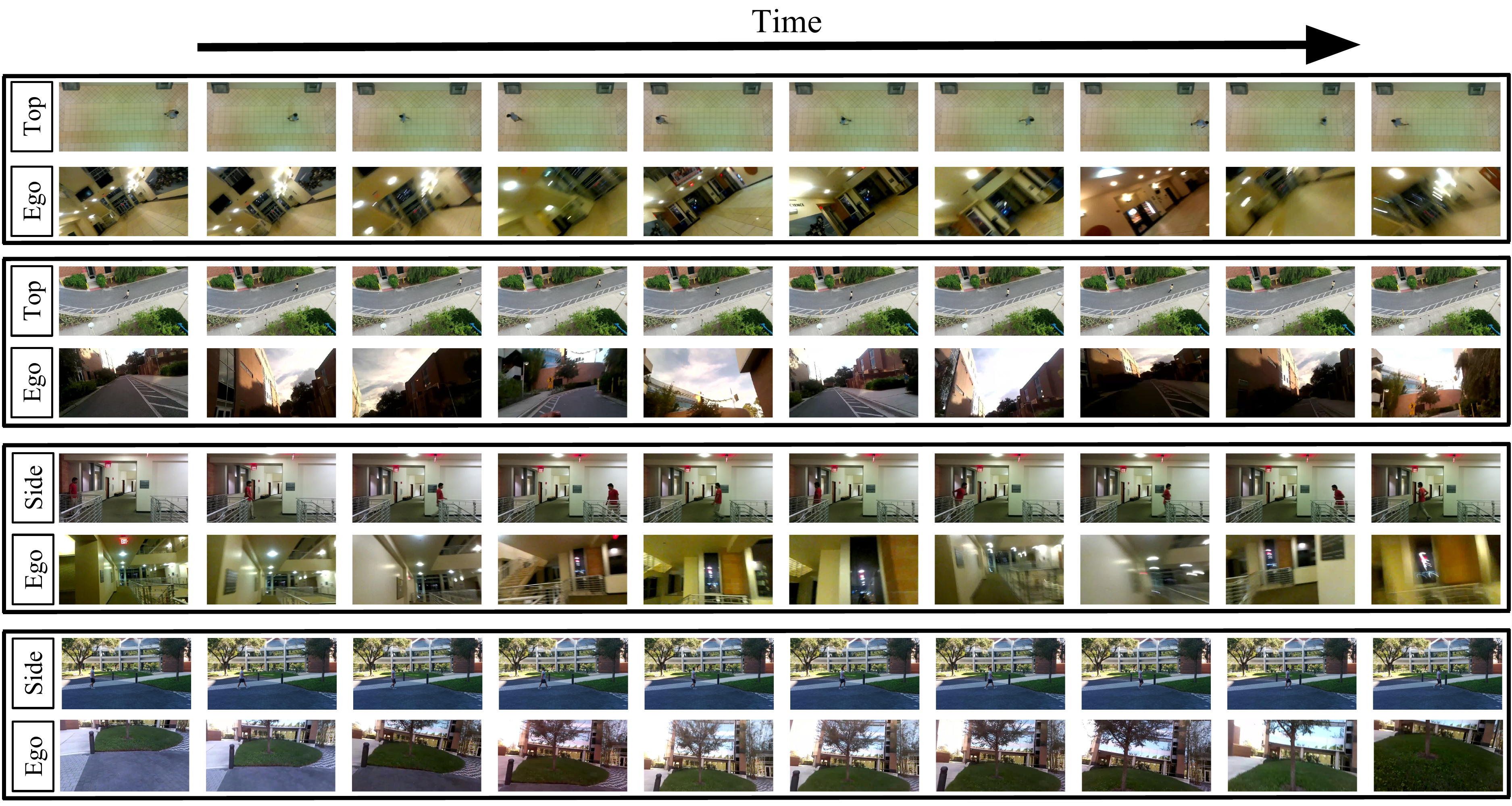}
\caption{A few examples of video pairs in our collected dataset. Each pair contains one egocentric and one exocentric video, simultaneously recorded by, and from an actor performing actions such as walking, jogging, running, hand waving, hand clapping, boxing and doing push-ups in order to capture a wide variety of motions.}
\label{fig:dataset_examples}
\end{figure*}

\subsection{Dataset}
To the best of our knowledge, there is no dataset containing simultaneously recorded egocentric and exocentric (side and top view videos) with a wide range of first and third person motions. Therefore, we collect a dataset containing 420 video pairs. Each video pair contains one egocentric and one exocentric (side or top-view) video. The pair of videos are temporally aligned, which will cause temporal features to correspond to each other. Some examples are shown in figure \ref{fig:dataset_examples}. Each pair was collected by asking an actor to perform a range of actions (walking, jogging, running, hand waving, hand clapping, boxing, push ups) covering a broad range of motions in front of an exocentric camera (top or side view), while wearing an egocentric body-worn camera capturing the actor's motion from the first person perspective. Each pair of videos is divided into pairs of temporally aligned short clips (16 frames each), and feature descriptors are extracted from each clip. Details about the number of videos and features used for training and testing are included in the following table. In order to increase the number of training and testing examples, we also add flipped versions of egocentric and side view videos, and also 12 rotated versions of the top-view videos (corresponding to 30 degrees of rotation).

\begin{table*}
\begin{center}
  \begin{tabular}{ l | c | c | c | c | c | c | c | c | }
 & \multicolumn{2}{c|}{Training Pairs} & \multicolumn{2}{c|}{Validation Pairs} & \multicolumn{2}{c|}{Testing Pairs} & \multicolumn{2}{c|}{Total Number of Pairs}  \\ 
  \cline{2-9}
     & $\#$Vid & $\#$Feat & $\#$Vid & $\#$Feat & $\#$Vid & $\#$Feat  & $\#$Vid & $\#$Feat \\ \hline    
    Ego-Side & 144 & 22,500 & 16 & 2,560 & 50 & 8,000& 210 & 33,060 \\ \hline
    Ego-Top & 144 & 69,120 & 16 & 7,680 & 50 & 24,000 & 210 & 100,800 \\
    \hline
  \end{tabular}  
\end{center}
\caption{Details of our dataset in terms of the number of training, validation and testing video and feature pairs.}
\label{tab:trainTest}
\end{table*}

\subsection{Evaluation}
The test set consists of a set of paired videos, from the source and target view. For each feature in the source view, we attempt to find its correct pair in the target view. To do so, we transform its feature to the target space using our learned mapping models. For each test feature, we evaluate the matching performance for each of the linear and non-linear mapping methods in terms of ranking. We then compute the area under the curve of the cumulative matching curve to have a quantitative measure of performance. We evaluate the mappings for two different features, C3D and HOOF, and four different scenarios including 1) egocentric to top-view, 2) top-view to egocentric, 3) egocentric to side-view, and 4) side-view to egocentric. 

\subsubsection{HOOF Features} 
\begin{figure*}[h]
\begin{center}
\begin{subfigure}{0.46\textwidth}
\includegraphics[width=1\linewidth]{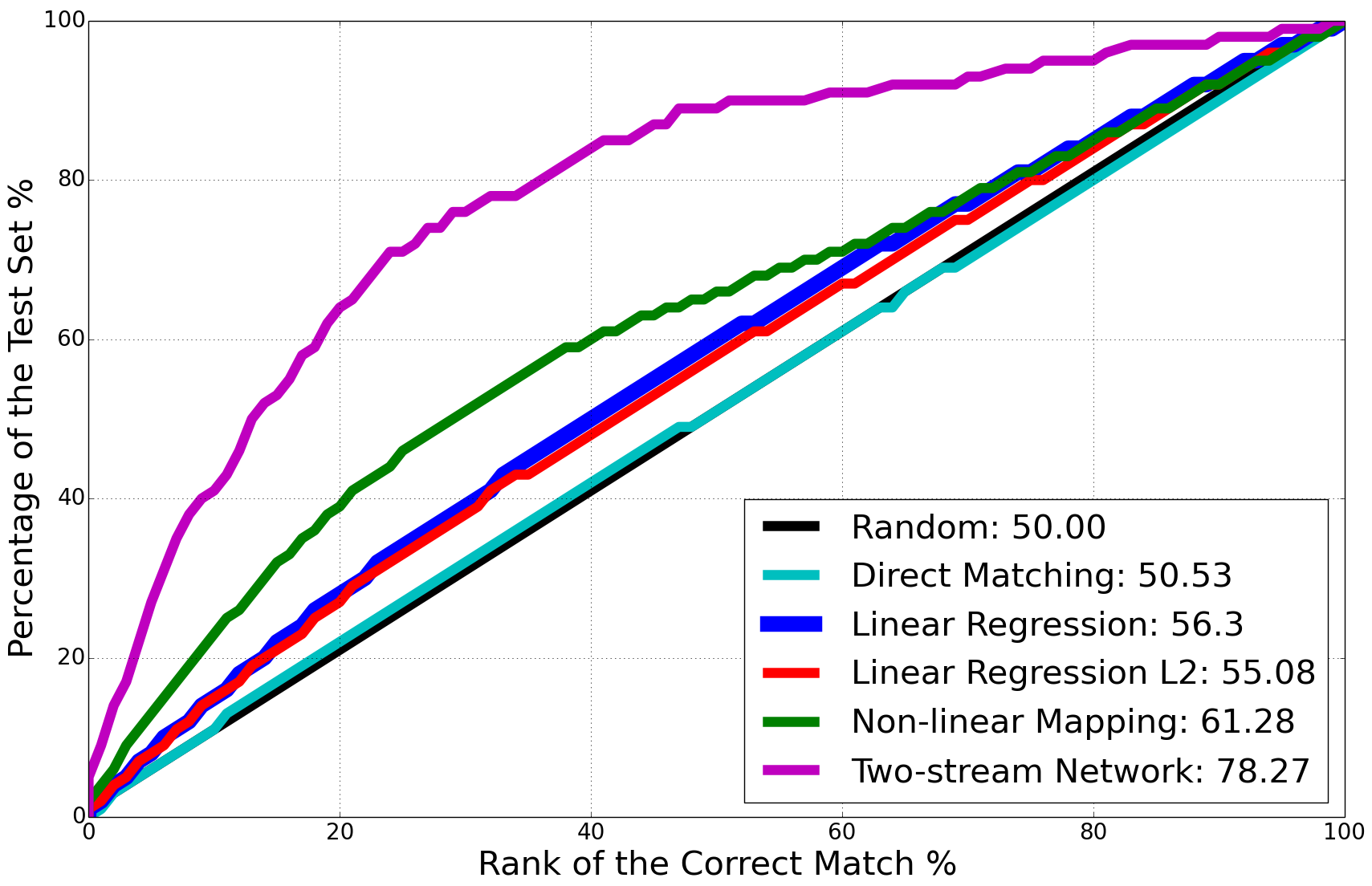}
\caption{HOOF top2ego}
\label{fig:cmc_HOOF_top2ego}
\end{subfigure}\hfill
\begin{subfigure}{0.46\textwidth}
\includegraphics[width=1\linewidth]{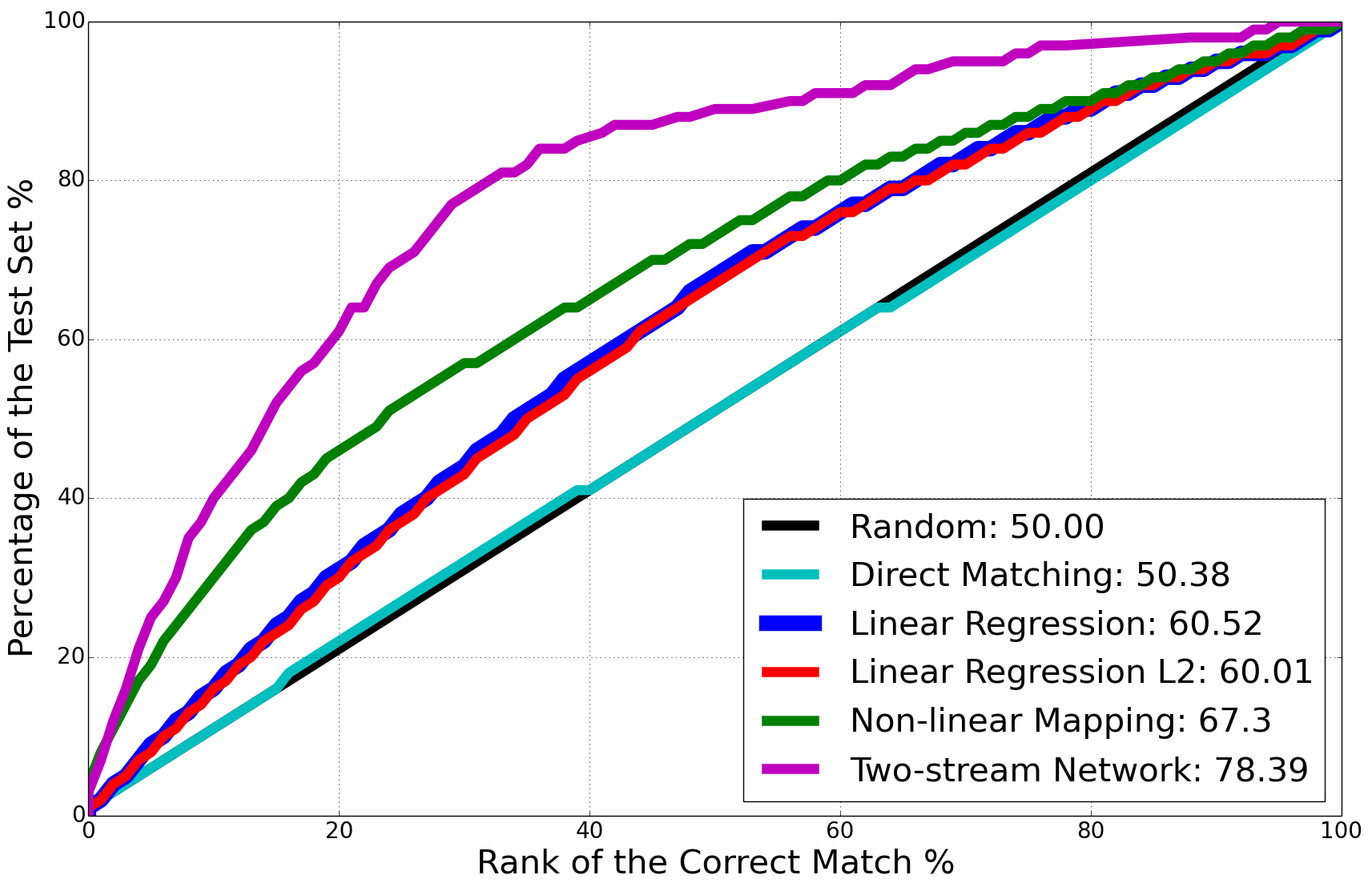}
\caption{HOOF ego2top}
\label{fig:cmc_HOOF_ego2top}
\end{subfigure}
\vspace{-8pt}
\caption{Mapping from egocentric view to top-view (left), and top-view to egocentric view (right) using HOOF features. As illustrated in the figures, in both cases linear regression and l2 regularized linear regression perform better than random. In both cases, our non-linear models drastically outperform the linear methods. The two stream classification network achieves the highest AUC. }
\label{fig:egoTopHOOF}
\end{center}
\vspace{-8pt}
\end{figure*}

\begin{figure*}[h]
\begin{center}
\begin{subfigure}{0.46\textwidth}
\includegraphics[width=1\linewidth]{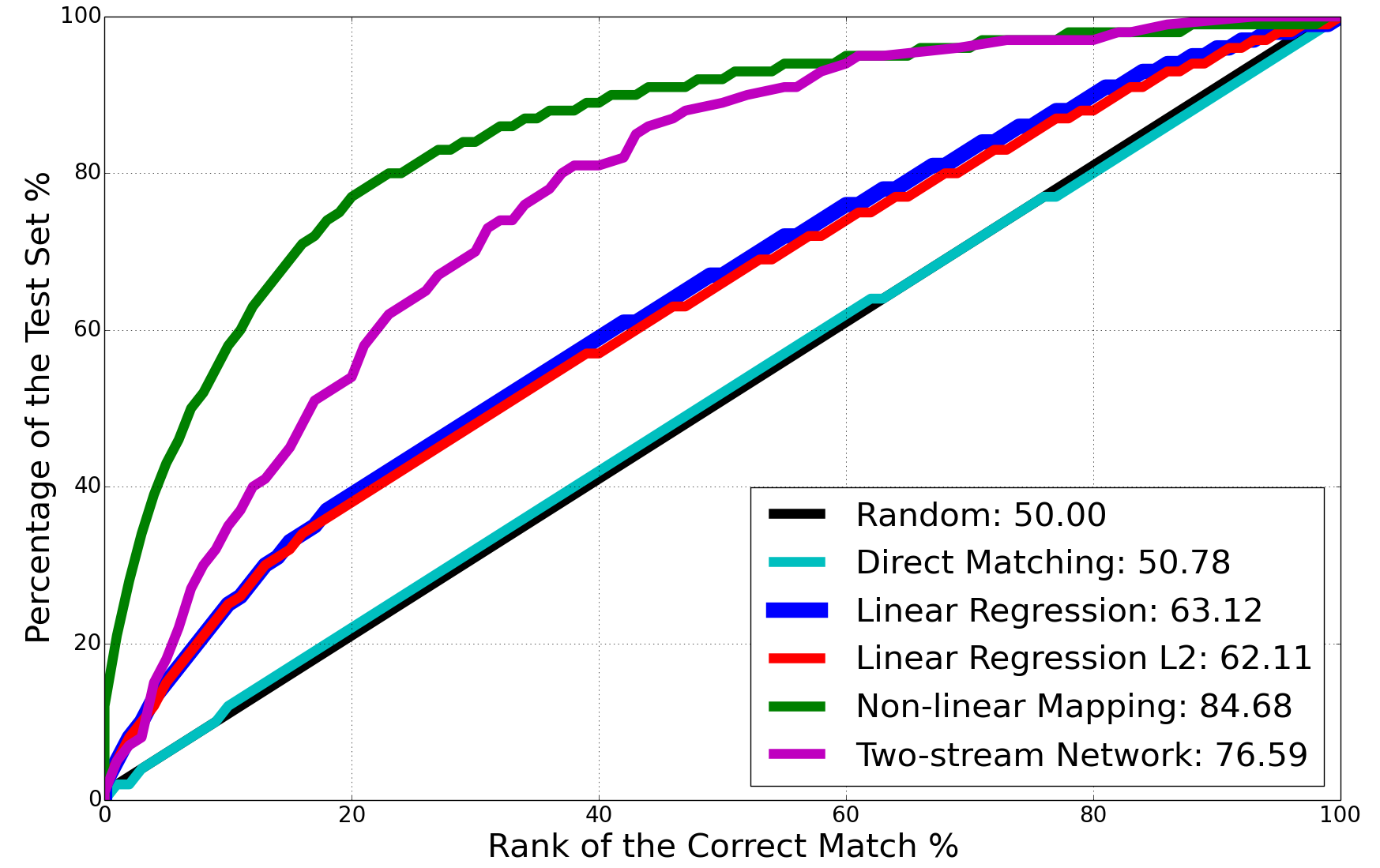}
\caption{HOOF ego2side}
\label{fig:features_1D}
\end{subfigure}\hfill
\begin{subfigure}{0.46\textwidth}
\includegraphics[width=1\linewidth]{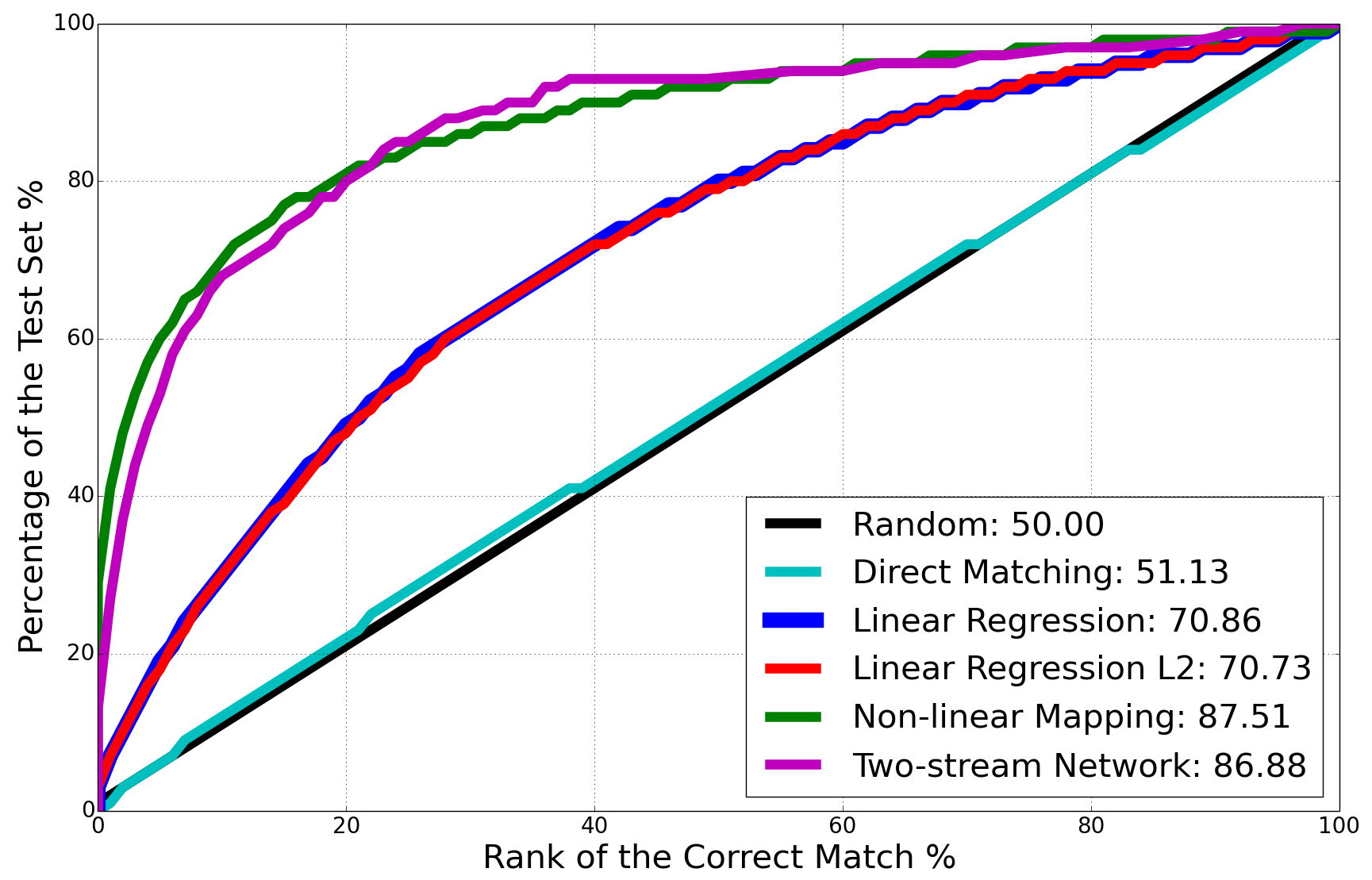}
\caption{HOOF side2ego}
\label{fig:cmc_HOOF_side}
\end{subfigure}
\vspace{-8pt}
\caption{Mapping from egocentric view to side-view (left), and side-view to egocentric view (right) using HOOF features. In both cases linear regression and l2 regularized linear regression perform better than random. The non-linear models drastically outperform the linear methods. However, in this case, the non-linear mapping model, outperforms the two stream classification network.}
\label{fig:egoSideHOOF}
\end{center}
\end{figure*}

As shown in figures \ref{fig:egoTopHOOF} and \ref{fig:egoSideHOOF}, in transforming the features from ego to top-view and vice versa, the non-linear models drastically outperform the linear models. In particular, the two-stream classification network achieves the highest accuracy. Further, for transferring features across egocentric and side-view, the non-linear mapping method outperforms the two-stream classification network. A summary of all the quantitative results with C3D features can be found in table \ref{tab:HOOF_all}.\\

\noindent\textbf{Mapping between Egocentric and Top-View:}
Figure \ref{fig:egoTopHOOF} shows the cumulative matching curves for mapping from egocentric to top-view and vice versa. In both scenarios, all models outperform chance, which confirms the possibility of finding a mapping between the two spaces. Non-linear methods perform more favorably compared to linear models. The two-stream classification based network achieves the highest accuracy.\\ 

\noindent\textbf{Mapping between Egocentric and Side-View:}
Figure \ref{fig:egoSideHOOF} shows the cumulative matching curves for mapping from egocentric to top-view and top-view to egocentric. In both scenarios, the linear mappings outperform chance, but under-perform the non-linear models. Also, the non-linear mapping with reconstruction objective achieves the best performance.\\ 

\begin{table*}
\begin{center}
\label{trainTest}
  \begin{tabular}{ | c | c | c | c | c | c | c | c | }
    \hline
    - & Random & Uniform & Regression & Regression L2 & Non-linear Mapping & Two-stream\\ \hline\hline
    Ego-Side & 50 & 50.78 & 63.12 & 62.11 & \textbf{84.68} & 76.59\\ \hline
    Side-Ego & 50 & 51.13 & 70.86 & 70.73 & \textbf{87.51} & 86.88\\ \hline
    Ego-Top & 50 & 50.53 & 56.3 & 55.08 & 61.28 & \textbf{78.27}\\ \hline
	Top-Ego & 50 & 50.38 & 60.52 & 60.01 & 67.3 & \textbf{78.39}\\ \hline
  \end{tabular}
  \end{center}
  \vspace{-8pt}
  \caption{Performance of different mapping methods on HOOF features. The non linear mapping with reconstruction objective achieves the most favorable result for mapping between egocentric and side view, and the two-stream network outperforms the rest in mapping between egocentric and top-view. Mapping egocentric and Side view generally achieves higher accuracy, compared to egocentric and top-view. This is intuitively justifiable given the fact that top-view has a more drastic difference with egocentric , compared to side view.}
  \label{tab:HOOF_all}
  \vspace{-8pt}
\end{table*}


\vspace{-10pt}
\subsubsection{C3D Features}
\begin{figure*}[h]
\begin{center}
\begin{subfigure}{0.46\textwidth}
\includegraphics[width=1\linewidth]{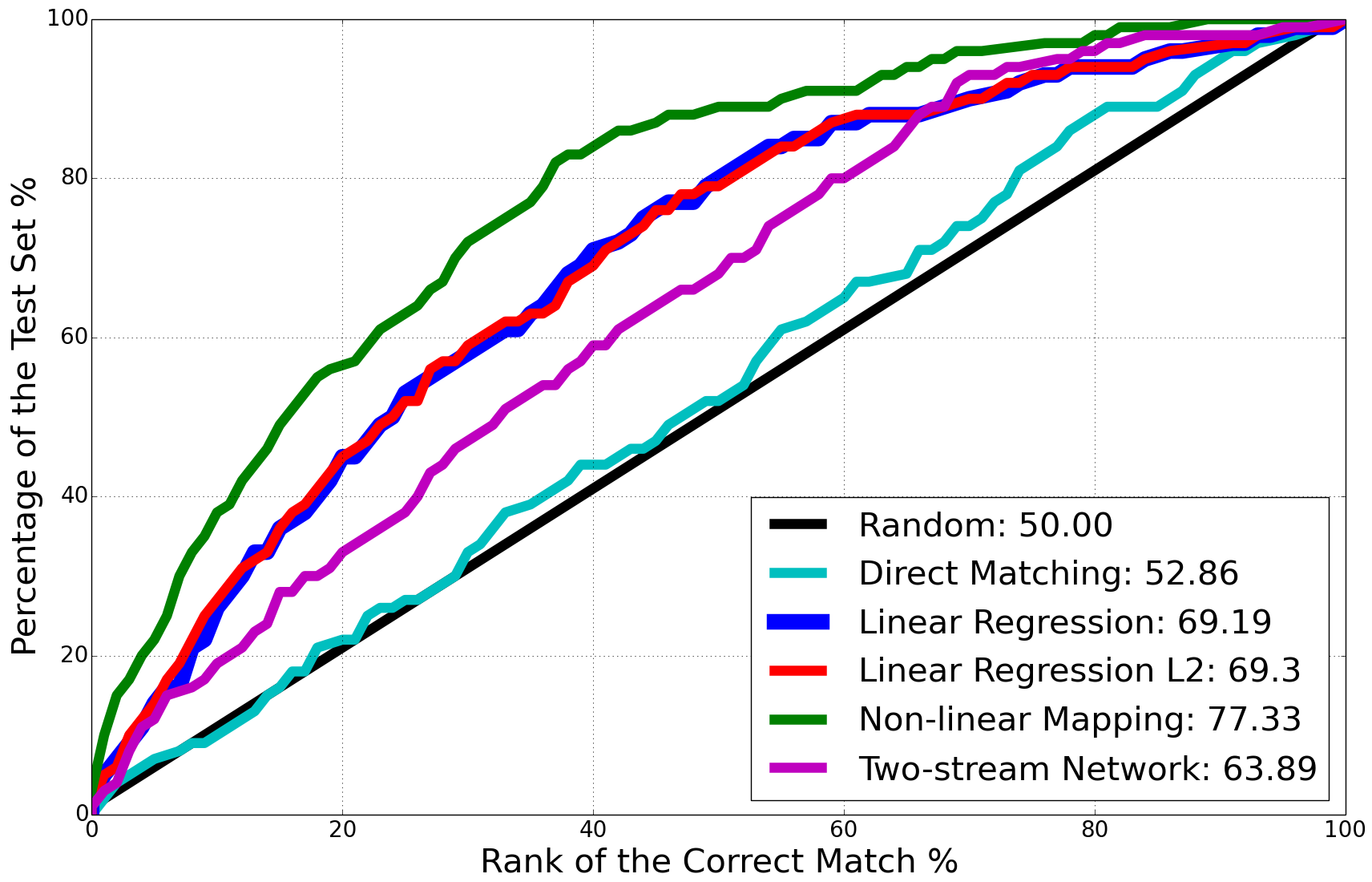}
\caption{C3D ego2top}
\label{fig:ego2top2_C3D_16}
\end{subfigure}\hfill
\begin{subfigure}{0.46\textwidth}
\centering
\includegraphics[width=1\linewidth]{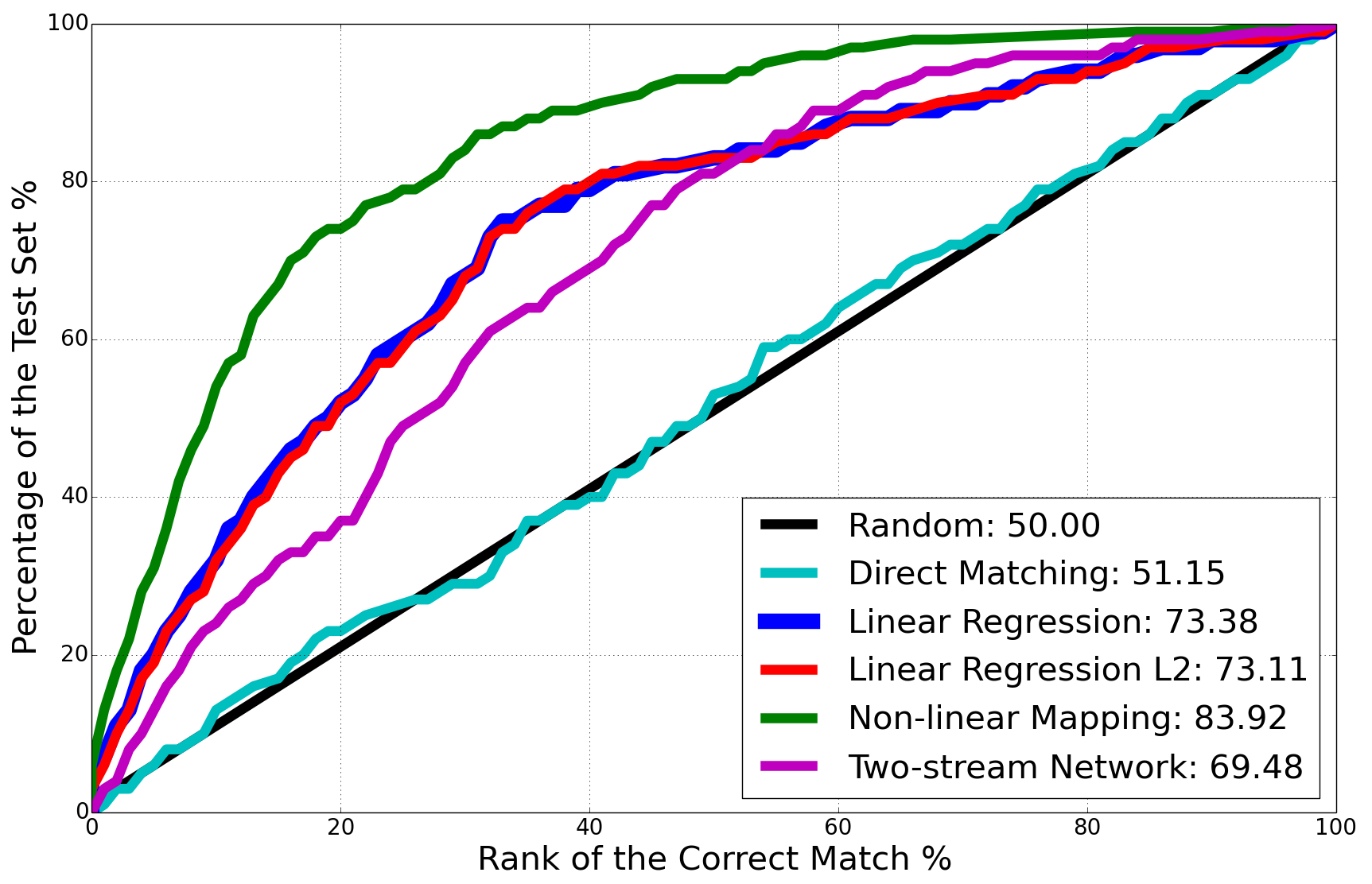}
\caption{C3D top2ego}
\label{fig:top2ego_C3D_16}
\end{subfigure}
\vspace{-8pt}
\caption{Mapping from egocentric view to top-view (left), and top-view to egocentric view (right) using C3D features. In both cases, linear regression and l2 regularized linear regression perform better than random. The non-linear mapping, outperforms the linear models. The two-stream classification network does not perform better than the linear models.}
\label{fig:egoTopC3D}
\end{center}
\end{figure*}

\begin{figure*}[h]
\begin{center}
\begin{subfigure}{0.46\textwidth}
\includegraphics[width=1\linewidth]{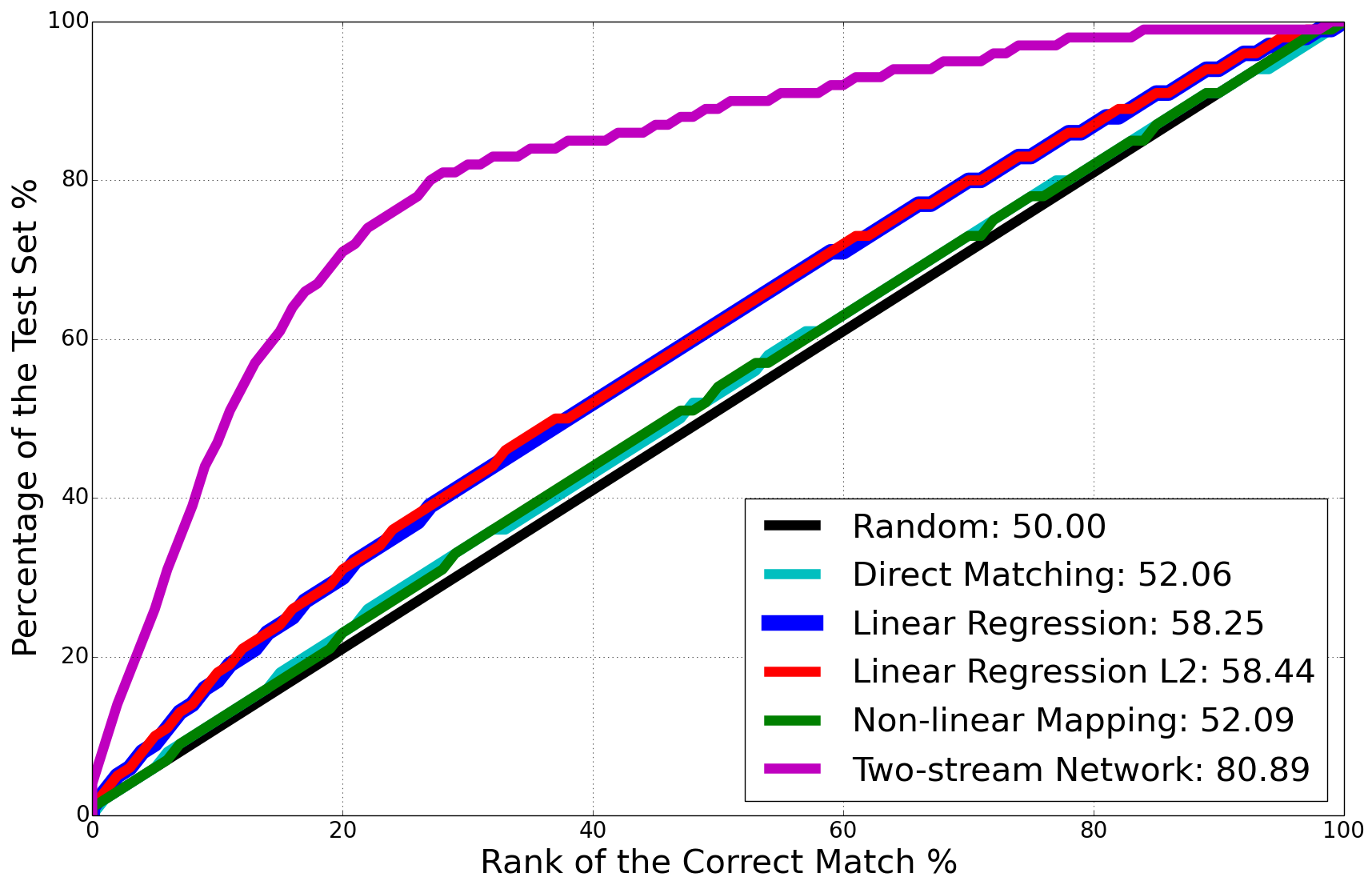}
\caption{C3D ego2side}
\label{fig:cmc_HOOF_side}
\end{subfigure}\hfill
\begin{subfigure}{0.46\textwidth}
\centering
\includegraphics[width=1\linewidth]{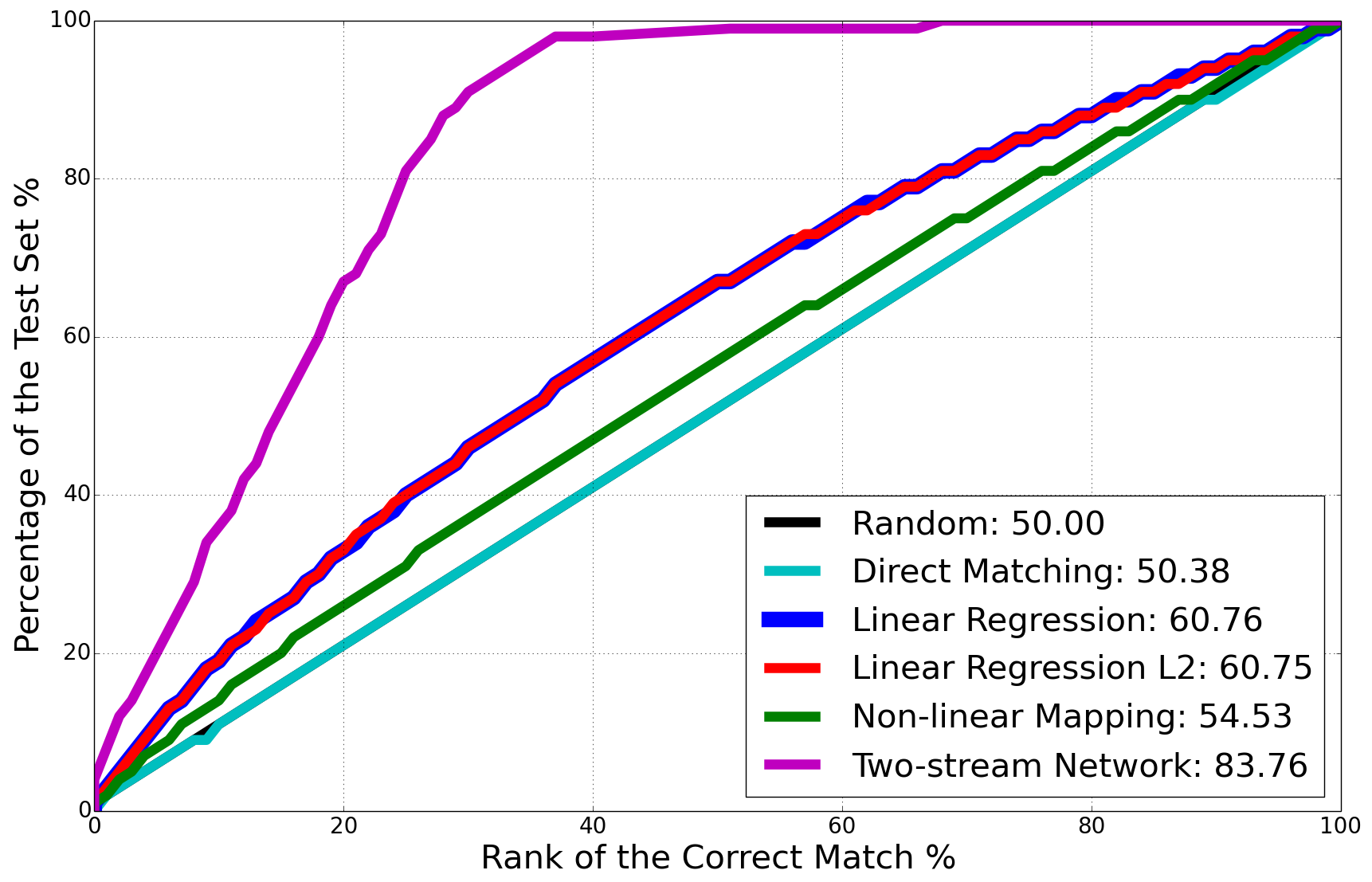}
\caption{C3D side2ego}
\label{fig:cmc_HOOF_side}
\end{subfigure}
\caption{Mapping from egocentric view to side-view (left), and side-view to egocentric view (right) using C3D features. In both cases linear regression and l2 regularized linear regression perform better than random.}
\label{fig:egoSideC3D}
\end{center}
\end{figure*}

Here, we evaluate the retrieval performance of our mapping methods using C3D features. Figures \ref{fig:egoTopC3D} and \ref{fig:egoSideC3D} show the retrieval performance for transforming features across ego to top and ego to side, respectively. 

It can be observed from figure \ref{fig:egoTopC3D} that the non-linear mapping outperforms linear mapping, while the two-stream classification model does not. On the other hand, in figure \ref{fig:egoSideC3D} where the aim is to transform features between egocentric and side view, the non-linear mapping does not perform as well as the linear methods. Here, the two-stream network drastically outperforms both. Generally, it can be seen that using C3D features, the non-linear models do not consistently outperform the linear models. We believe this is because C3D features, extracted from the last fully connected layers of the 3DCNNS, offer fully meaningful and independent information. This makes them more suitable for linear models. As a result, not every non-linear model should necessarily be able to outperform a linear model when trained on those features. According to this, one may expect non-linear models to estimate an identity transformation. However, \cite{he2015deep} shows that it is very hard for a non-linear fully connected layer to learn an identity transformation accurately. Therefore, additional non-linearity can further reduce the accuracy of a mapping of a feature which is already tailored to work with a linear model such as C3D. A summary of all the quantitative results with C3D features can be found in table \ref{tab:C3D_all}.\\

\noindent\textbf{Mapping between Egocentric and Top-View:} Figure \ref{fig:egoTopC3D} shows the cumulative matching curves for mapping from egocentric to top-view and top-view to egocentric. It can be seen that in both scenarios, the non-linear mapping achieves the best performance.\\ 

\noindent\textbf{Mapping between Egocentric and Side-View:}
Figure \ref{fig:egoSideC3D} illustrates the performance for mapping from egocentric to side-view and vice versa. Here, we find that the two-stream classification network achieves the best performance. 

\begin{table*}
\begin{center}
\label{trainTest}
  \begin{tabular}{ | c | c | c | c | c | c | c | c | }
    \hline
    - & Random & Uniform & Regression & Regression L2 & Non-linear Mapping & Two-stream\\ \hline\hline
    Ego-Side & 50 & 52.06 & 58.25 & 58.44 & 52.09 & \textbf{80.89}\\ \hline
    Side-Ego & 50 & 50.38 & 60.76 & 60.75 & 54.53 & \textbf{83.76}\\ \hline
    Ego-Top & 50 & 52.86 & 69.19 & 69.3 & \textbf{77.33} & 63.89\\ 
\hline
	Top-Ego & 50 & 51.15 & 73.38 & 73.11 & \textbf{83.92} & 69.48\\ \hline
  \end{tabular}
  \end{center}
  \vspace{-8pt}
  \caption{Performance of different mapping methods on C3D features. The non-linear mapping with reconstruction objective gives the most favorable results for mapping between egocentric and top-view, and the two-stream network outperforms the rest in mapping between egocentric and side-view.}
\label{tab:C3D_all}
\end{table*}


\section{Conclusion and Discussion}
Inspired by the mirror neuron concept, we explored the possibility of transforming motion information across two drastically different views: egocentric (or first-person) and exocentric (or third-person). We showed that it is possible to learn a transformation, linear and non-linear, across the two spaces. We observe that depending on the scenario (e.g., ego to top, side to ego) and the feature type, linear models can outperform non-linear models. The opposite can happen as well. Overall, using both HOOF and C3D features, transferring motion to and from side view leads to higher accuracy compared to top-view. This intuitively makes sense, since side views are often visually more similar and contain more information regarding an activity.

For future, we plan to extend our work for action recognition and visual tracking. For the former, a classifier will be trained from data of a source domain, using the proposed features, and will be applied to the data of a target domain. For the latter, we will use the recorded egocentric video, offline or online, to better track a person in top-view surveillance camera. We will consider using more sophisticated spatio-temporal features as well as domain adaptation and task transfer approaches.

In this work, we explored the possibility of transferring knowledge across egocentric and exocentric domains. We believe that our work can be a stepping stone to further explore the relationship between the two domains, with possible applications in action recognition, and identification.

\clearpage

{\small
\bibliographystyle{ieee}
\bibliography{Thesis_bib}
}

\end{document}